%% file: emnlp2020.tex
\documentclass[11pt,a4paper]{article}
\usepackage[hyperref]{emnlp2020}
\usepackage{times}
\usepackage{latexsym}
\usepackage{amssymb}
\usepackage{mwe}
\usepackage{subcaption}
\usepackage{amsmath}
\usepackage{booktabs}
\usepackage{tabularx}
\usepackage{multirow}
\usepackage{algorithm}
\usepackage{algpseudocode}
\usepackage{soul}
\usepackage{xcolor}

\hypersetup{
    colorlinks,
    linkcolor={red!50!black},
    citecolor={blue!50!black},
    urlcolor={blue!80!black}
}

\graphicspath{{figures/}}

\usepackage{letltxmacro}
\LetLtxMacro\oldttfamily\ttfamily
\DeclareRobustCommand{\ttfamily}{\oldttfamily\csname ttsize\endcsname}
\newcommand{\setttsize}[1]{\def\ttsize{#1}}%

\algnewcommand{\IIf}[1]{\State\algorithmicif\ #1\ \algorithmicthen}
\algnewcommand{\EndIIf}{\unskip\ \algorithmicend\ \algorithmicif}

\usepackage{microtype}

\aclfinalcopy 
\def\aclpaperid{1117} 

\title{\papertitle}

\author{Victor Zhong \\
  \small{University of Washington} \\
  \small{Seattle, WA} \\
  \texttt{vzhong@cs.washington.edu} \\\And
  Mike Lewis \\
  \small{Facebook AI Research} \\
  \small{Seattle, WA} \\
  \texttt{mikelewis@fb.com} \\\And
  Sida I. Wang \\
  \small{Facebook AI Research} \\
  \small{Seattle, WA} \\
  \texttt{sidawang@fb.com} \\\And
  Luke Zettlemoyer \\
  \small{University of Washington} \\
  \small{Facebook AI Research} \\
  \small{Seattle, WA} \\
  \texttt{lsz@cs.washington.edu}
}

\date{}

\begin{document}
\include{victor}
\setttsize{\small}

\maketitle
\begin{abstract}
We propose~\papertitle~(\modelnameshort) to adapt an existing semantic parser to new environments (e.g.~new database schemas). 
\modelnameshort~combines a forward semantic parser with a backward utterance generator to synthesize data (e.g.~utterances and SQL queries) in the new environment, then selects cycle-consistent examples to adapt the parser.
Unlike data-augmentation, which typically synthesizes unverified examples in the training environment,~\modelnameshort~synthesizes examples in the new environment whose input-output consistency are verified.
On the Spider, Sparc, and CoSQL zero-shot semantic parsing tasks,~\modelnameshort~improves logical form and execution accuracy of the baseline parser.
Our analyses show that~\modelnameshort~outperforms data-augmentation in the training environment, performance increases with the amount of~\modelnameshort-synthesized data, and cycle-consistency is central to successful adaptation.
\end{abstract}

\section{Introduction}

Semantic parsers~\citep{zelle1996LearningTP,zettlemoyer2005LearningTM,Liang2011LearningDC} build executable meaning representations for a range of tasks such as question-answering~\citep{yih2014semantic}, robotic control~\citep{matuszek2013Learning}, and intelligent tutoring systems~\citep{graesser2005autotutor}.
However, they are usually engineered for each application environment.
For example, a language-to-SQL parser trained on an university management database struggles when deployed to a sales database.
How do we adapt a semantic parser to new environments where no training data exists?

\begin{figure*}[t]
    \centering
    \includegraphics[width=0.9\linewidth]{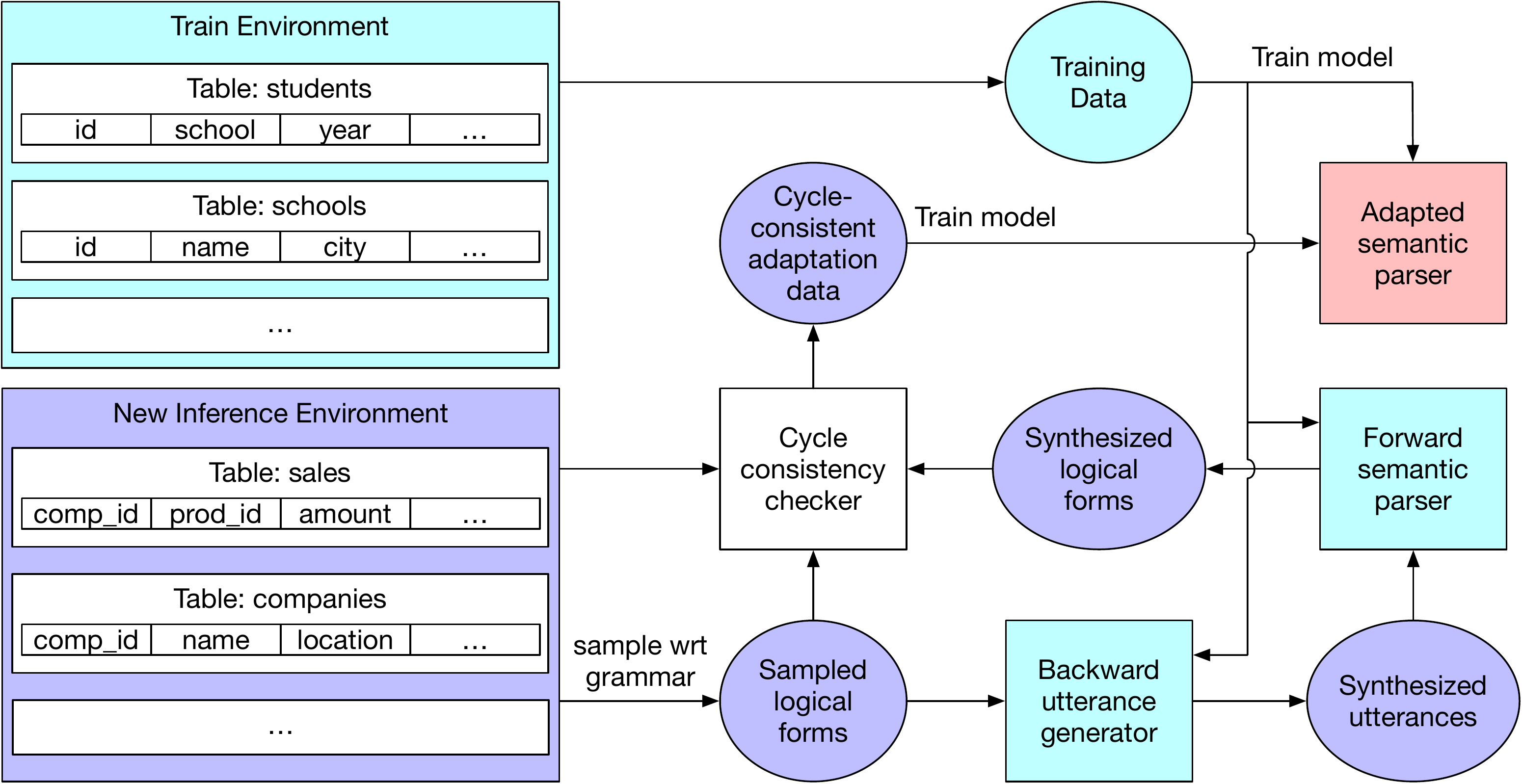}
    \caption{
    \papertitle.
    \modelnameshort~adapts a parser to new inference environments.
    Data and models for training and inference environments are respectively shown in blue and purple.
    Output is shown in red.
    First, we train a parser and a utterance generator using training data.
    We then sample logical forms in the inference environment and generate corresponding utterances.
    We parse the generated utterances and check for cycle-consistency between the parse and the sampled logical form~(see Section~\ref{sec:cycle}).
    Consistent pairs of utterance and logical form are used to adapt the parser to the inference environment.
    }
    \label{fig:main}
    \vspace{-0.15in}
\end{figure*}

We propose~\modelnameemphasized
, which adapts existing semantic parsers to new environments by synthesizing new, cycle-consistent data.
In the previous example,~\modelnameshort~synthesizes high-quality sales questions and SQL queries using the new sales database, then adapts the parser using the synthesized data.
This procedure is shown in Figure~\ref{fig:main}.
\modelnameshort~is complementary to prior modeling work in that it can be applied to any model architecture, in any domain where one can enforce cycle-consistency by evaluating equivalence between logical forms.
Compared to data-augmentation, which typically synthesizes unverified data in the training environment,~\modelnameshort~instead synthesizes consistency-verified data in the new environment.

\modelnameshort~synthesizes data for consistency-verified adaptation using a forward semantic parser and a backward utterance generator.
Given a new environment (e.g.~new database), we first sample logical forms with respect to a grammar (e.g.~SQL grammar conditioned on new database schema).
Next, we generate utterances corresponding to these logical forms using the generator.
Then, we parse the generated utterances using the parser, keeping those whose parses are equivalent to the original sampled logical form (more in Section~\ref{sec:cycle}).
Finally, we adapt the parser to the new environment by training on the combination of the original data and the synthesized cycle-consistent data.

We evaluate~\modelnameshort~on the Spider, Sparc, and CoSQL~\citep{yu2018spider,yu2019cosql,yu2019sparc} language-to-SQL zero-shot semantic parsing tasks which test on unseen databases.
\modelnameshort~improves logical form and execution accuracy of the baseline parser on all tasks, successfully adapting the existing parser to new environments.
In further analyses, we show that~\modelnameshort~outperforms data augmentation in the training environment.
Moreover, adaptation performance increases with the amount of~\modelnameshort-synthesized data.
Finally, we show that cycle-consistency is critical to synthesizing high-quality examples in the new environment, which in turn allows for successful adaptation and performance improvement.\footnote{We open-sourced this project at \url{https://github.com/vzhong/gazp}.}

\section{\papertitle}
\label{sec:method}
 
Semantic parsing involves producing a \textbf{logical form} $\logicalform$ that corresponds to an input \textbf{utterance} $\utterance$, such that executing $\logicalform$ in the \textbf{environment} $\environment$ produces the desired \textbf{denotation} $\exe{\logicalform}{\environment}$.
In the context of language-to-SQL parsing, $\logicalform$ and $\environment$ correspond to SQL queries and databases.

We propose~\modelnameshort~for zero-shot semantic parsing, where inference environments have not been observed during training (e.g.~producing SQL queries in new databases).
\modelnameshort~consists of a \textbf{forward semantic parser} $\parser(\utterance, \environment) \rightarrow \logicalform$, which produces a logical form $\logicalform$ given an utterance $\utterance$ in environment $\environment$, and a \textbf{backward utterance generator} $\generator(\logicalform, \environment) \rightarrow \utterance$.
The models $\parser$ and $\generator$ condition on the environment by reading an \textbf{environment description} $\envdesc$, which consists of a set of \textbf{documents} $\envdoc$.
In the context of SQL parsing, the description is the database schema, which consists of a set of table schemas (i.e.~documents).

We assume that the logical form consists of three types of tokens: \textbf{syntax candidates} $\cand_\syn$ from a fixed syntax vocabulary (e.g. SQL syntax), \textbf{environment candidates} $\cand_\environment$ from the environment description (e.g. table names from database schema), and \textbf{utterance candidates} $\cand_\utterance$ from the utterance (e.g. values in SQL query).
Finally, $\cand_\environment$ tokens have corresponding spans in the description $\envdoc$.
For example, a SQL query $\logicalform$ consists of columns $\cand_\environment$ that directly map to related column schema (e.g.~table, name, type) in the database schema $\envdesc$.

In~\modelnameshort~, we first train the forward semantic parser $\parser$ and a backward utterance generator $\generator$ in the training environment $\environment$.
Given a new inference environment $\infenvironment$, we sample logical forms $\logicalform$ from $\infenvironment$ using a grammar.
For each $\logicalform$, we generate a corresponding utterance $\gutterance = \generator(\logicalform, \infenvironment)$.
We then parse the generated utterance into a logical form $\glogicalform = \parser(\gutterance, \infenvironment)$.
We combine cycle-consistent examples from the new environment, for which $\glogicalform$is equivalent to $\logicalform$, with the original labeled data to retrain and adapt the parser.
Figure~\ref{fig:main}~illustrates the components of~\modelnameshort.
We now detail the sampling procedure, forward parser, backward generator, and cycle-consistency.

\begin{algorithm}[t]
\caption{Query sampling procedure.}
\label{alg:sample}
\begin{algorithmic}[1]
  \small
  \State $d \gets \sfunction{UniformSample}(\mathit{AllDBs})$
  \State $\Coarse \gets \emptyset$
  \For{$\coarse \in \mathit{CoarseTemplates}$}
    \IIf{$d.\sfunction{CanFill}(\coarse)$}
      $\Coarse.\sfunction{Add}(\coarse)$
    \EndIIf
  \EndFor
  \State $\coarse^\prime \gets \sfunction{Sample}(P_\Coarse)$
  \State $d^\prime \gets d.\sfunction{randAssignColsToSlots}()$
  \For{$s \in \coarse^\prime.\sfunction{colSlots}()$}
    \State $c \gets d^\prime.\sfunction{getCol}(s)$
    \State $\coarse^\prime.\sfunction{replSlotWithCol}(s, c)$
  \EndFor
  \For{$s \in \coarse^\prime.\sfunction{valSlots}()$}
    \State $c \gets d^\prime.\sfunction{getCol}(s)$
    \State $v \gets c.\sfunction{uniformSampleVals}()$
    \State $\coarse^\prime.\sfunction{replSlotWithVal}(s, v)$
  \EndFor
  
  \Comment{Return $\coarse^\prime$}
\end{algorithmic}
\end{algorithm}

\subsection{Query sampling}
\label{sec:sampling}

To synthesize data for adaptation, we first sample logical forms $\logicalform$ with respect to a grammar.
We begin by building an empirical distribution over $\logicalform$ using the training data.
For language-to-SQL parsing, we preprocess queries similar to~\citet{zhang2019editing}~and further replace mentions of columns and values with typed slots to form \textbf{coarse templates} $\Coarse$.
For example, the query~\texttt{SELECT T1.id, T2.name FROM Students AS T1 JOIN Schools AS T2 WHERE T1.school = T2.id AND T2.name = 'Highland Secondary'}, after processing, becomes~\texttt{SELECT key1, text1 WHERE text2 = val}.
Note that we remove~\texttt{JOIN}s which are later filled back deterministically after sampling the columns.
Next, we build an empirical distribution $P_\Coarse$ over these coarse templates by counting occurrences in the training data.
The sampling procedure is shown in Algorithm~\ref{alg:sample} for the language-to-SQL example.
Invalid queries and those that execute to the empty set are discarded.

Given some coarse template $\coarse =$~\texttt{SELECT key1, text1 WHERE text2 = val}, the function $d.\sfunction{CanFill}(\coarse)$ returns whether the database $d$ contains sufficient numbers of columns.
In this case, at the minimum, $d$ should have a key column and two text columns.
The function $d.\sfunction{randAssignColsToSlots()}$ returns a database copy $d^\prime$ such that each of its columns is mapped to some identifier $\texttt{text1}$, $\texttt{key1}$ etc.

Appendix~\ref{app:sampling}~quantifies query coverage of the sampling procedure on the Spider task, and shows how to extend Algorithm~\ref{alg:sample} to multi-turn queries.

\begin{figure*}[t]
    \centering
    \includegraphics[width=0.9\linewidth]{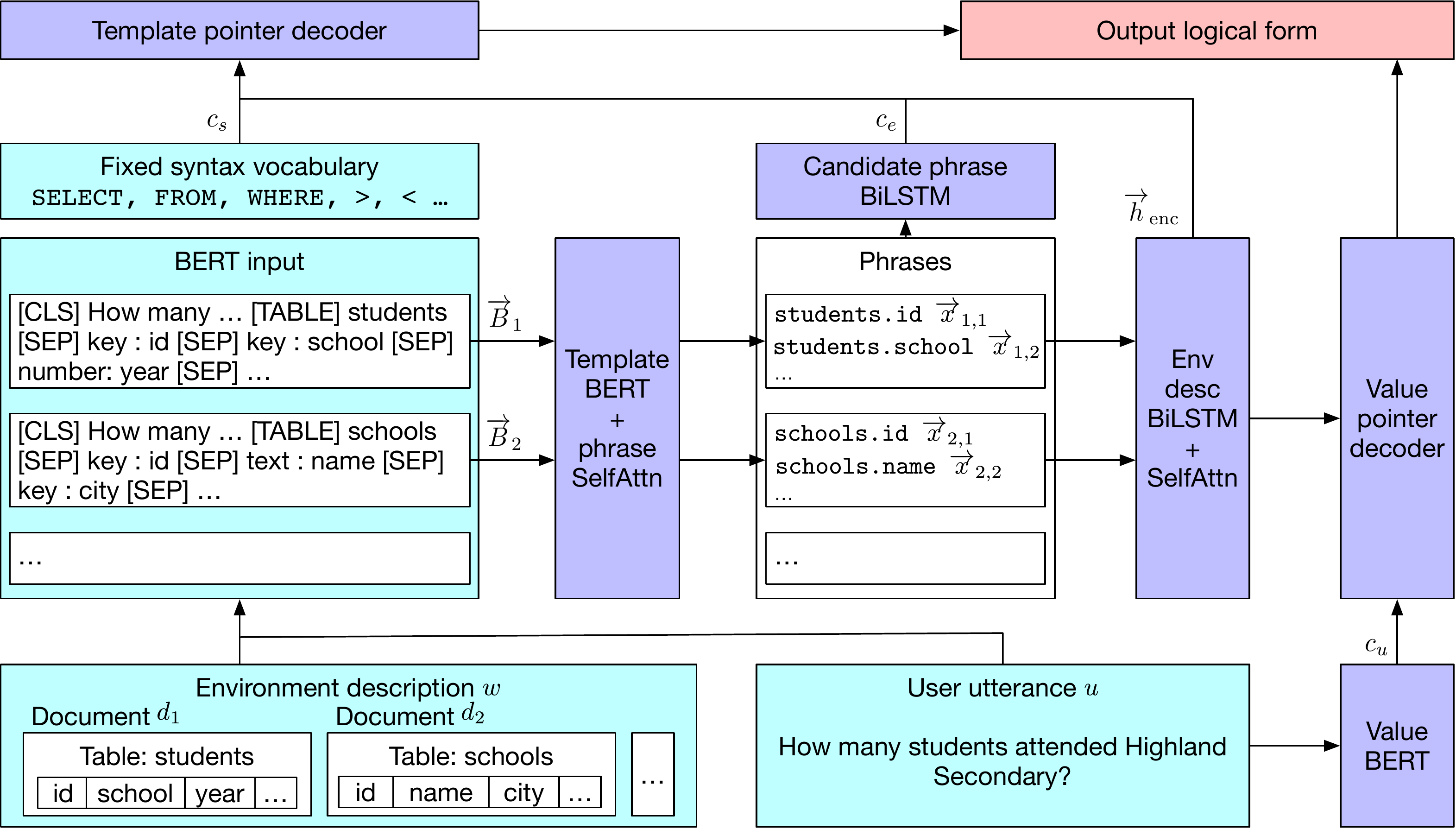}
    \caption{
    Forward semantic parser.
    Model components are shown in purple, inputs in blue, and outputs in red.
    First, we cross-encode each environment description text and the utterance using BERT.
    We then extract document-level phrase representations for candidate phrases in each text, which we subsequently encode using LSTMs to form input and environment-level candidate phrase representations.
    A pointer-decoder attends over the input and selects among candidates to produce the output logical form.
    }
    \vspace{-0.15in}
    \label{fig:forward}
\end{figure*}

\begin{figure*}[t]
    \centering
    \includegraphics[width=0.8\linewidth]{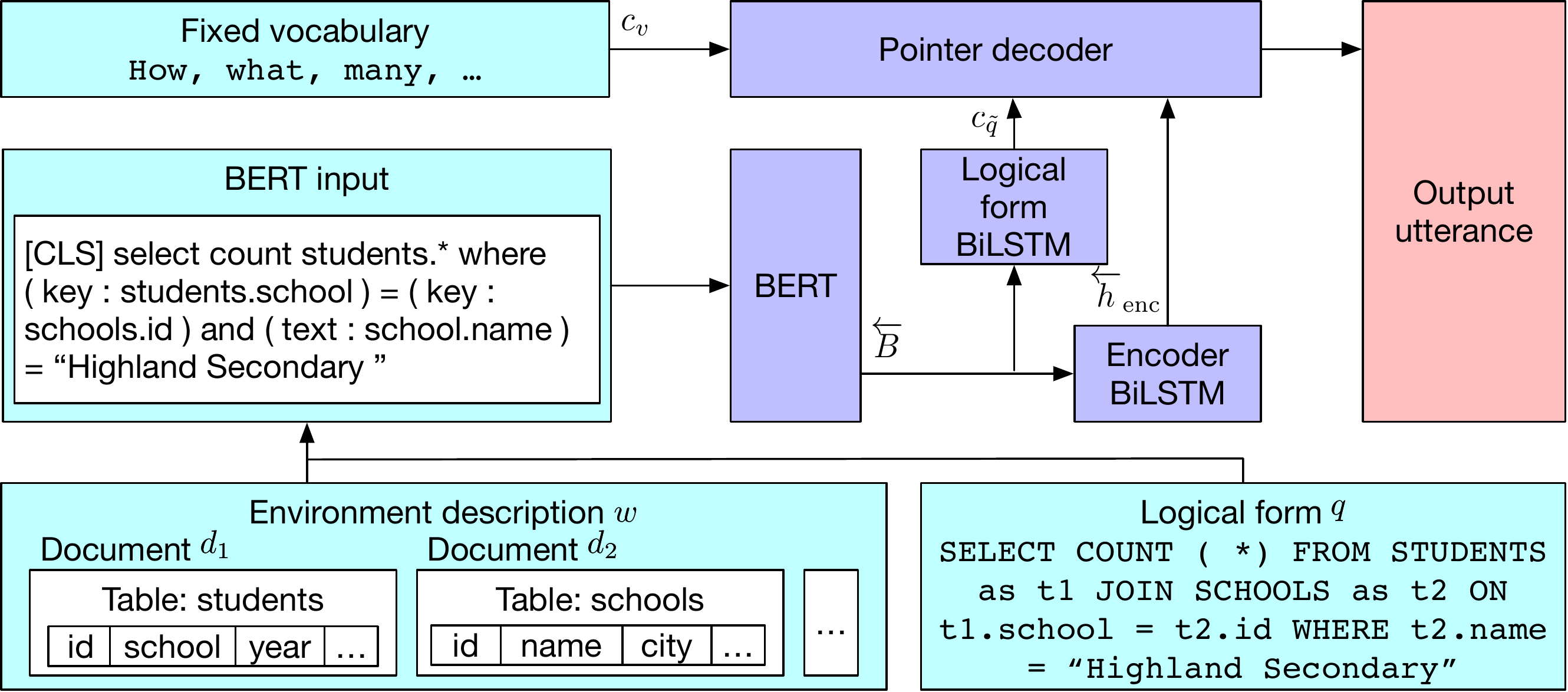}
    \caption{
    Backward utterance generator.
    Model components are shown in purple, inputs in blue, and outputs in red.
    First, we encode the input logical form along with environment description for each of its symbols.
    we subsequently encode using LSTMs to form the input and environment-level candidate token representations.
    A pointer-decoder attends over the input and selects among candidate representations to produce the output utterance.
    }
    \vspace{-0.15in}
    \label{fig:backward}
\end{figure*}

\subsection{Forward semantic parser}
\label{sec:parser}

The forward semantic parser $\parser$ produces a logical form $\logicalform = \parser(\utterance, \environment)$ for an utterance $\utterance$ in the environment $\environment$.
We begin by cross-encoding $\utterance$ with the environment description $\envdesc$ to model coreferences.
Since $\envdesc$ may be very long (e.g.~entire database schema), we instead cross-encode $\utterance$ with each document $\envdoc_i$ in the description (e.g.~each table schema) similar to~\citet{zhang2019editing}.
We then combine each environment candidate $\cand_{\environment, i}$ across documents (e.g. table columns) using RNNs, such that the final representations capture dependencies between $\cand_\environment$ from different documents.
To produce the logical form $\logicalform$, we first generate a logical form template $\template$ whose utterance candidates $\cand_\utterance$ (e.g.~SQL values) are replaced by slots.
We generate $\template$ with a pointer-decoder that selects among syntax candidates $\cand_\syn$ (e.g. SQL keywords) and environment candidate $\cand_\environment$ (e.g. table columns).
Then, we fill in slots in $\template$ with a separate decoder that selects among $\cand_\utterance$ in the utterance to form $\logicalform$.
Note that logical form template $\template$ is distinct from coarse templates $\coarse$ described in sampling (Section~\ref{sec:sampling}).
Figure~\ref{fig:forward} describes the forward semantic parser.

Let $\utterance$ denote words in the utterance, and $\envdoc_i$ denote words in the $i$th document in the environment description.
Let $\concat{a; b}$ denote the concatenation of $a$ and $b$.
First, we cross-encode the utterance and the document using BERT~\citep{devlin2018bert}, which has led to improvements on a number of NLP tasks.
\begin{eqnarray}
\forwardbertenc_i &= \bertforward(\concat{\utterance; \envdoc_i})
\label{eq:bertforward}
\end{eqnarray}
Next, we extract environment candidates in document $i$ using self-attention.
Let $s$, $e$ denote the start and end positions of the $j$th environment candidate in the $i$th document.
We compute an intermediate representation $\phrase_{ij}$ for each environment candidate:
\begin{eqnarray}
a &=& \softmax(W [\forwardbertenc_{is}; ... \forwardbertenc_{ie}] + b)\\
\phrase_{ij} &=& \sum_{k=s}^{e} a_k \forwardbertenc_{ik}
\end{eqnarray}
For ease of exposition, we abbreviate the above self-attention function as $\phrase_{ij} = \selfattn(\forwardbertenc_i[s:e])$
Because $\phrase_{ij}$ do not model dependencies between different documents, we further process $\phrase$ with bidirectional LSTMs~\citep{Hochreiter1997LongSM}.
We use one LSTM followed by self-attention to summarize each $i$th document:
\begin{eqnarray}
\forward{\state}_{\enc, i} = \selfattn(\BiLSTM(\concat{\phrase_{i1}; \phrase_{i2}; ...}))
\end{eqnarray}
We use another LSTM to build representations for each environment candidate $\cand_{\environment, i}$
\begin{eqnarray}
\cand_\environment = \BiLSTM(\concat{x_{11}; x_{12}; ... x_{21}; x_{22} ...})
\end{eqnarray}
We do not share weights between different LSTMs and between different self-attentions.

Next, we use a pointer-decoder~\citep{vinyals-pointer_networks} to produce the output logical form template $\template$ by selecting among a set of candidates that corresponds to the union of environment candidates $\cand_\environment$ and syntax candidates $\cand_\syn$.
Here, we represent a syntax token using its BERT word embedding.
The representation for all candidate representations $\forwardcand$ is then obtained as
\begin{eqnarray}
\forwardcand = \concat{\cand_{\environment, 1}; \cand_{\environment,2}; ... \cand_{\syn, 1};  \cand_{\syn, 2}; ...}
\end{eqnarray}
At each step $t$ of the decoder, we first update the states of the decoder LSTM:
\begin{eqnarray}
\state_{\dec, t} = \lstm(\forward{\cand}_{\template_{t-1}}, \state_{\dec, t-1})
\end{eqnarray}
Finally, we attend over the document representations given the current decoder state using dot-product attention~\citep{bahdanau-align_and_translate}:
\begin{eqnarray}
\hat{a}_t &=& \softmax( \state_{\dec, t} \forward{\state}_{\enc} ^\intercal )\\
v_t &=& \sum_i \hat{a}_{t, i} \forward{\state}_{\enc, i}
\end{eqnarray}
The score for the $i$th candidate $\forwardcand_i$ is
\begin{eqnarray}
o_t &=& \hat{W} \concat{\state_{\dec, t}; v_t} + \hat{b}\\
s_{t, i} &=& o_t \forwardcand_i^\intercal\\
\template_t &=& \argmax(s_{t})
\end{eqnarray}

\paragraph{Value-generation.}
The pervious template decoder produces logical form template $\template$, which is not executable because it does not include utterance candidates $\cand_\utterance$.
To generate full-specified executable logical forms $\logicalform$, we use a separate value pointer-decoder that selects among utterance tokens.
The attention input for this decoder is identical to that of the template decoder.
The pointer candidates $\cand_\utterance$ are obtained by running a separate BERT encoder on the utterance $\utterance$.
The produced values are inserted into each slot in $\template$ to form $\logicalform$.

Both template and value decoders are trained using cross-entropy loss with respect to the ground-truth sequence of candidates.

\subsection{Backward utterance generator}
\label{sec:generator}

The utterance generator $\generator$ produces an utterance $\utterance = \generator(\logicalform, \environment)$ for the logical form $\logicalform$ in the environment $\environment$.
The alignment problem between $\logicalform$ and the environment description $\envdesc$ is simpler than that between $\utterance$ and $\envdesc$ because environment candidates $\cand_\environment$ (e.g.~column names) in $\logicalform$ are described by corresponding spans in $\envdesc$ (e.g.~column schemas in database schema).
To leverage this deterministic alignment, we augment $\cand_\environment$ in $\logicalform$ with relevant spans from $\envdesc$, and encode this augmented logical form $\auglogicalform$.
The pointer-decoder selects among words $\cand_\vocab$ from a fixed vocabulary (e.g. when, where, who) and words $\cand_\auglogicalform$ from $\auglogicalform$.
Figure~\ref{fig:backward}~illustrates the backward utterance generator.

First, we encode the logical form using BERT.
\begin{eqnarray}
\backward{\bertenc} = \bertbackward(\auglogicalform)
\end{eqnarray}
Next, we apply a bidirectional LSTM to obtain the input encoding $\backward{\state}_\enc$ and another bidirectional LSTM to obtain representations of tokens in the augmented logical form $\cand_\auglogicalform$.
\begin{eqnarray}
\backward{\state}_\enc &=& {\BiLSTM}(\backward{\bertenc})\\
\cand_\auglogicalform &=& {\BiLSTM}(\backward{\bertenc})
\end{eqnarray}
To represent $\cand_\vocab$, we use word embeddings from $\bertbackward$.
Finally, we apply a pointer-decoder that attends over $\backward{\state}_\enc$ and selects among candidates $\backwardcand = \concat{\cand_\auglogicalform ; \cand_\vocab}$ to obtain the predicted utterance.

\subsection{Synthesizing cycle-consistent examples}
\label{sec:cycle}

Having trained a forward semantic parser $\parser$ and a backward utterance generator $\generator$ in environment $\environment$, we can synthesize new examples with which to adapt the parser in the new environment $\infenvironment$.
First, we sample a logical form $\logicalform$ using a grammar (Algorithm~\ref{alg:sample} in Section~\ref{sec:sampling}).
Next, we predict an utterance $\gutterance = \generator(\logicalform, \infenvironment)$.
Because $\generator$ was trained only on $\environment$, many of its outputs are low-quality or do not correspond to its input $\logicalform$.
On their own, these examples ($\gutterance, \logicalform$) do not facilitate parser adaptation (see Section~\ref{sec:results} for analyses).

To filter out low-quality examples, we additionally predict a logical form $\glogicalform = \parser(\gutterance, \infenvironment)$, and keep only examples that are \textbf{cycle consistent} --- the synthesized logical form $\glogicalform$ is equivalent to the originally sampled logical form $\logicalform$ in $\infenvironment$.
In the case of SQL parsing, the example is cycle-consistent if executing the synthesized query $\exe{\glogicalform}{\infenvironment}$ results in the same denotation (i.e.~same set of database records) as executing the original sampled query $\exe{\logicalform}{\infenvironment}$.
Finally, we combine cycle-consistent examples synthesized in $\infenvironment$ with the original training data in $\environment$ to retrain and adapt the parser.

\section{Experiments}
\label{sec:experiments}

\begin{figure*}[t]
\centering
\begin{subfigure}[b]{0.95\linewidth}
   \includegraphics[width=\linewidth]{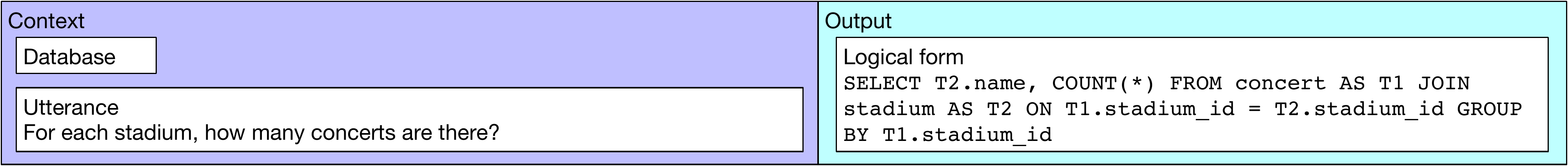}
   \caption{Example from Spider.}
   \label{fig:spider}
\end{subfigure}
\begin{subfigure}[b]{0.95\linewidth}
   \includegraphics[width=\linewidth]{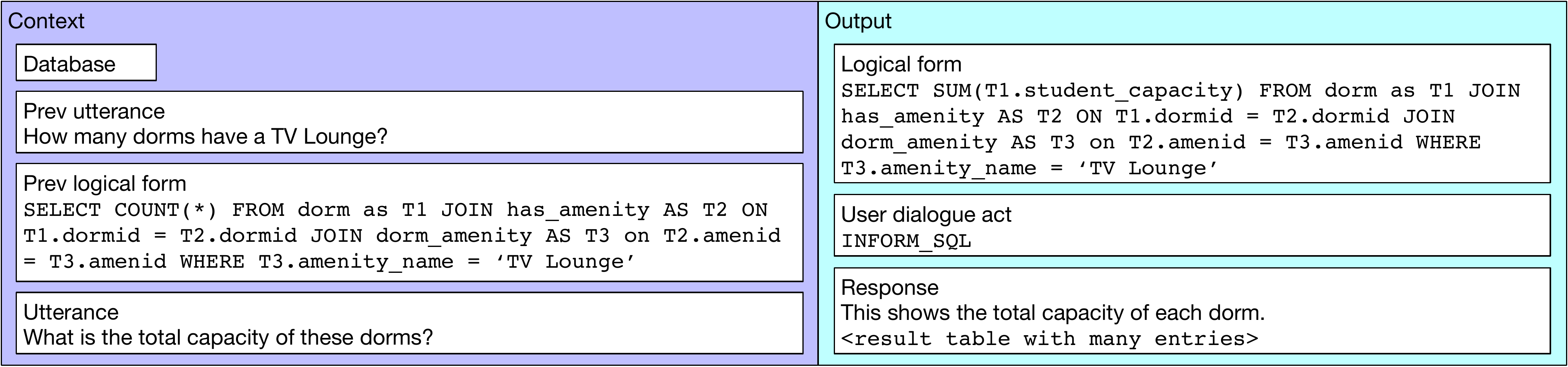}
   \caption{Example from CoSQL.}
   \label{fig:cosql}
\end{subfigure}
\vspace{-0.15in}
\caption{
Examples from (a) Spider and (b) CoSQL.
Context and output are respectively shown in purple and blue.
We do not show Sparc because its data format is similar to CoSQL, but without user dialogue act prediction and without response generation.
For our experiments, we produce the output logical form given the data, utterance, and the previous logical form if applicable.
During evaluation, the previous logical form is the output of the model during the previous turn (i.e. no teacher forcing on ground-truth previous output).
}
\vspace{-0.1in}
\label{fig:datasets}
\end{figure*}

We evaluate performance on the Spider~\citep{yu2018spider}, Sparc~\citep{yu2019sparc}, and CoSQL~\citep{yu2019cosql} zero-shot semantic parsing tasks.
Table~\ref{tab:datasets}~shows dataset statistics.
Figure~\ref{fig:datasets}~shows examples from each dataset.
For all three datasets, we use preprocessing steps from~\citet{zhang2019editing} to preprocess SQL logical forms.
Evaluation consists of \textbf{exact match over logical form templates} (EM) in which values are stripped out, as well as \textbf{execution accuracy} (EX).
Official evaluations also recently incorporated \textbf{fuzz-test accuracy} (FX) as tighter variant of execution accuracy.
In fuzz-testing, the query is executed over randomized database content numerous times.
Compared to an execution match, a fuzz-test execution match is less likely to be spurious (e.g.~the predicted query coincidentally executes to the correct result).
FX implementation is not public as of writing, hence we only report test FX.

\begin{table}[t]
\centering
\begin{tabularx}{\linewidth}{llll}
\toprule
                 & Spider & Sparc  & CoSQL  \\ \midrule
\# database      & 200    & 200    & 200    \\
\# tables        & 1020   & 1020   & 1020   \\
\# utterances    & 10,181 & 4298   & 3007   \\
\# logical forms & 5,693  & 12,726 & 15,598 \\
multi-turn       & no     & yes    & yes    \\ \bottomrule
\end{tabularx}
\vspace{-0.12in}
\caption{Dataset statistics.}
\vspace{-0.10in}
\label{tab:datasets}
\vspace{-0.15in}
\end{table}

\paragraph{Spider.}
Spider is a collection of database-utterance-SQL query triplets.
The task involves producing the SQL query given the utterance and the database.
Figure~\ref{fig:forward}~and~\ref{fig:backward}~show preprocessed input for the parser and generator.

\paragraph{Sparc.}
In Sparc, the user repeatedly asks questions that must be converted to SQL queries by the system.
Compared to Spider, Sparc additionally contains prior interactions from the same user session (e.g. database-utterance-query-previous query quadruplets).
For Sparc evaluation, we concatenate the previous system-produced query (if present) to each utterance.
For example, suppose the system was previously asked ``where is Tesla born?'' and is now asked ``how many people are born there?'', we produce the utterance \texttt{[PREV] SELECT birth\_place FROM people WHERE name = 'Tesla' [UTT] how many people are born there ?}
For training and data synthesis, the ground-truth previous query is used as generation context for forward parsing and backward utterance generation.

\paragraph{CoSQL.}
CoSQL is combines task-oriented dialogue and semantic parsing.
It consists of a number of tasks, such as response generation, user act prediction, and state-tracking.
We focus on state-tracking, in which the user intent is mapped to a SQL query.
Similar to~\citet{zhang2019editing}, we restrict the context to be the previous query and the current utterance.
Hence, the input utterance and environment description are obtained in the same way as that used for Sparc.

\begin{table*}[t]
\centering
\addtolength{\tabcolsep}{-1.7pt}
\begin{tabular}{@{}llllllllllllllll@{}}
\toprule
\multirow{3}{*}{Model} & \multicolumn{5}{c}{Spider}       & \multicolumn{5}{c}{Sparc}        & \multicolumn{5}{c}{CoSQL}        \\ \cmidrule(l){2-6} \cmidrule(l){7-11} \cmidrule(l){12-16}
 & \multicolumn{2}{c}{dev} & \multicolumn{3}{c}{test} & \multicolumn{2}{c}{dev} & \multicolumn{3}{c}{test} & \multicolumn{2}{c}{dev} & \multicolumn{3}{c}{test} \\  \cmidrule(l){2-3} \cmidrule(l){4-6} \cmidrule(l){7-8} \cmidrule(l){9-11} \cmidrule(l){12-13} \cmidrule(l){14-16}
                       & EM   & EX   & EM   & EX   & FX   & EM   & EX   & EM   & EX   & FX   & EM   & EX   & EM   & EX   & FX   \\ \midrule
EditSQL                & 57.6 & n/a  & 53.4 & n/a  & n/a  & 47.2 & n/a  & 47.9 & n/a  & n/a  & 39.9 & n/a  & 40.8 & n/a  & n/a  \\ \midrule
Baseline               & 56.8 & 55.4 & 52.1 & 49.8 & 51.1 & 46.4 & 44.0 & 45.9 & 43.5 & 42.8 & 39.3 & 36.6 & 37.2 & 34.9 & 33.8 \\
GAZP                   & \textbf{59.1} & \textbf{59.2} & \textbf{53.3} & \textbf{53.5} & \textbf{51.7} & \textbf{48.9} & \textbf{47.8} & 45.9 & \textbf{44.6} & \textbf{43.9} & \textbf{42.0} & \textbf{38.8} & \textbf{39.7} & \textbf{35.9} & \textbf{36.3} \\ \bottomrule
\end{tabular}
\caption{
Development set evaluation results on Spider, Sparc, and CoSQL.
\textbf{EM} is exact match accuracy of logical form templates without values.
\textbf{EX} is execution accuracy of fully-specified logical forms with values.
\textbf{FX} is execution accuracy from fuzz-testing with randomized databases.
\textbf{Baseline} is the forward parser without adaptation.
\textbf{EditSQL} is a state-of-the-art language-to-SQL parser that produces logical form templates that are not executable.
}
\vspace{-0.1in}
\label{tab:results}
\end{table*}

\begin{table*}[t]
\centering
\begin{tabularx}{\textwidth}{Xlllllllll}
\toprule
\multirow{2}{*}{Model} & \multicolumn{3}{c}{Spider}      & \multicolumn{3}{c}{Sparc}       & \multicolumn{3}{c}{CoSQL}       \\ \cmidrule(lr){2-4} \cmidrule(lr){5-7} \cmidrule(lr){8-10}
                       & \multicolumn{1}{l}{EM} & \multicolumn{1}{l}{EX} & \# syn & \multicolumn{1}{l}{EM} & \multicolumn{1}{l}{EX} & \# syn & \multicolumn{1}{l}{EM} & \multicolumn{1}{l}{EX} & \# syn \\ \midrule
Baseline        & 56.8 & 55.4 & 40557 & 46.4 & 44.0 & 45221 & 39.3 & 36.6 & 33559 \\
\modelnameshort & 59.1 & \textbf{59.2} & 40557 & \textbf{48.9} & \textbf{47.8} & 45221 & \textbf{42.0} & \textbf{38.8} & 33559 \\ \midrule
nocycle         & 55.6 & 52.3 & 97655 & 41.1 & 40.0 & 81623 & 30.7 & 30.8 & 78428 \\
syntrain        & 54.8 & 52.1 & 39721 & 47.4 & 45.2 & 44294 & 38.7 & 34.3 & 31894 \\
EM consistency  & \textbf{61.6} & 56.9 & 35501 & 48.4 & 45.9 & 43521 & 41.9 & 37.7 & 31137 \\ \bottomrule
\end{tabularx}
\caption{
Ablation performance on development sets.
For each one, 100,000 examples are synthesized, out of which queries that do not execute or execute to the empty set are discarded.
``nocycle'' uses adaptation without cycle-consistency.
``syntrain'' uses data-augmentation on training environments.
``EM consistency'' enforces logical form instead of execution consistency.
}
\label{tab:ablation}
\vspace{-0.15in}
\end{table*}

\subsection{Results}

\label{sec:results}

We primarily compare~\modelnameshort~with the baseline forward semantic parser, because prior systems produce queries without values which are not executable.
We include one such non-executable model, EditSQL~\citep{zhang2019editing}, one of the top parsers on Spider at the time of writing, for reference.
However, EditSQL EM is not directly comparable because of different outputs.

Due to high variance from small datasets, we tune the forward parser and backward generator using cross-validation.
We then retrain the model with early stopping on the development set using hyperparameters found via cross-validation.
For each task, we synthesize 100k examples, of which $\sim$40k are kept after checking for cycle-consistency.
The adapted parser is trained using the same hyperparameters as the baseline.
Please see appendix~\ref{app:hyperparameters} for hyperparameter settings.
Appendix~\ref{app:synthesized} shows examples of synthesized adaptation examples and compares them to real examples.

Table~\ref{tab:results}~shows that adaptation by~\modelnameshort~results in consistent performance improvement across Spider, Sparc, and CoSQL in terms of EM, EX, and FX.
We also examine the performance breakdown across query classes and turns (details in appendix~\ref{app:breakdown}).
First, we divide queries into difficulty classes based on the number of SQL components, selections, and conditions~\citep{yu2018spider}.
For example, queries that contain more components such as \texttt{GROUP}, \texttt{ORDER}, \texttt{INTERSECT}, nested subqueries, column selections, and aggregators, etc are considered to be harder.
Second, we divide multi-turn queries into how many turns into the interaction they occur for Sparc and CoSQL~\citep{yu2019sparc,yu2019cosql}.
We observe that the gains in~\modelnameshort~are generally more pronounced in more difficult queries and in turns later in the interaction. 
Finally, we answer the following questions regarding the effectiveness of cycle-consistency and grounded adaptation.

\paragraph{Does adaptation on inference environment outperform data-augmentation on training environment?}
For this experiment, we synthesize data on training environments instead of inference environments.
The resulting data is similar to data augmentation with verification.
As shown in the ``syntrain'' row of Table~\ref{tab:ablation}, retraining the model on the combination of this data and the supervised data leads to overfitting in the training environments.
A method related to data-augmentation is jointly supervising the model using the training data in the reverse direction, for example by generating utterance from query~\citep{fried2018speaker,cao2019dual}.
For Spider, we find that this dual objective (57.2 EM) underperforms GAZP adaptation (59.1 EM).
Our results indicate that adaptation to the new environment significantly outperforms augmentation in the training environment.

\paragraph{How important is cycle-consistency?}
For this experiment, we do not check for cycle-consistency and instead keep all synthesized queries in the inference environments.
As shown in the ``nocycle'' row of Table~\ref{tab:ablation}, the inclusion of cycle-consistency effectively prunes $\sim$60\% of synthesized examples, which otherwise significantly degrade performance.
This shows that enforcing cycle-consistency is crucial to successful adaptation.

In another experiment, we keep examples that have consistent logical forms, as deemed by string match (e.g.~$\logicalform == \glogicalform$), instead of consistent denotation from execution.
The ``EM consistency'' row of Table~\ref{tab:ablation} shows that this variant of cycle-consistency also improves performance.
In particular, EM consistency performs similarly to execution consistency, albeit typically with lower execution accuracy.

\begin{figure}[t]
    \centering
    \includegraphics[width=\linewidth]{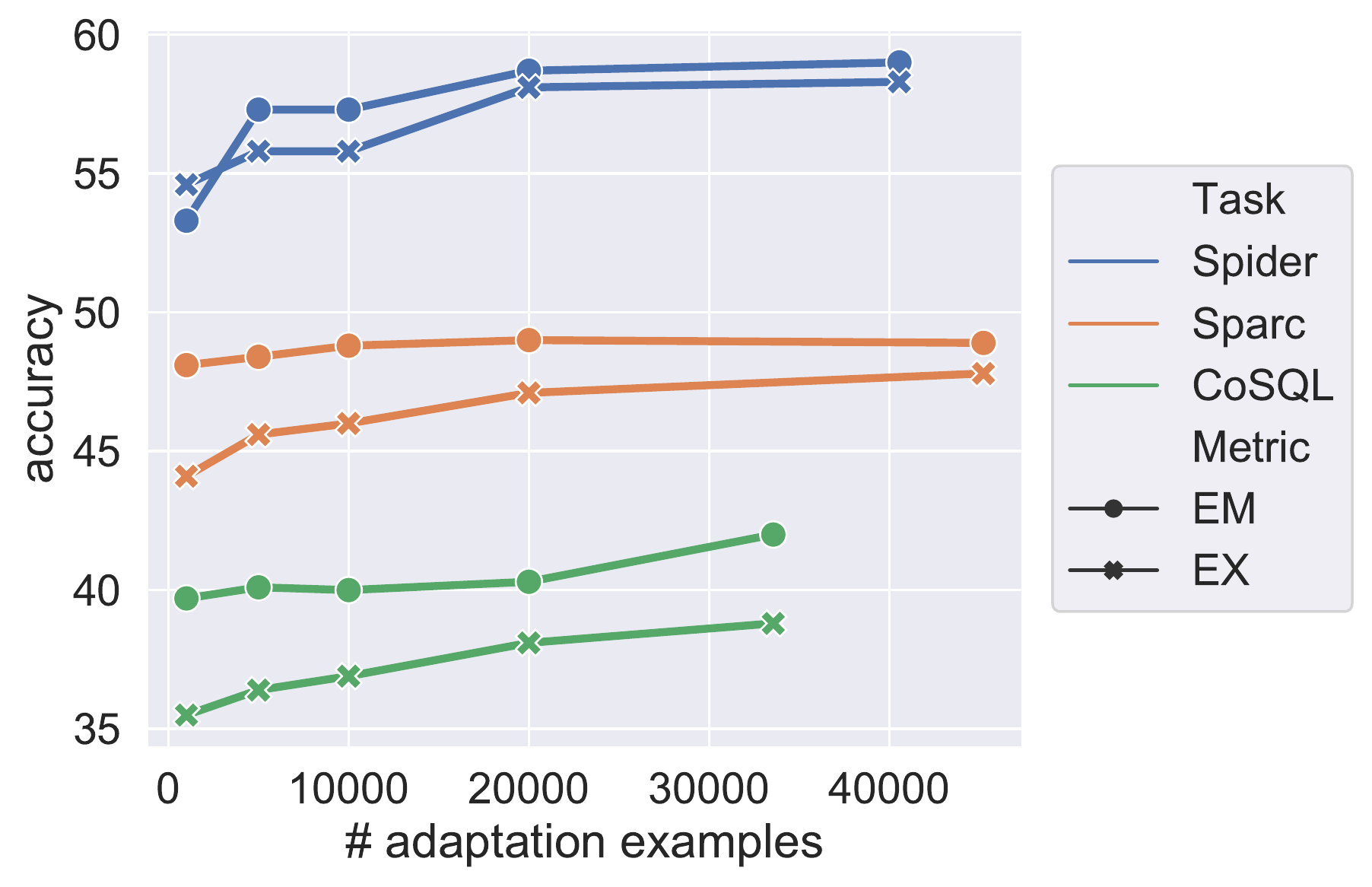}
    \vspace{-0.2in}
    \caption{
    Effect of amount of synthesized data on adaptation performance on the development set.
    EM and EX denote template exact match and logical form execution accuracy, respectively.
    The $x$-axis shows the number of cycle-consistent examples synthesized in the inference environments (e.g. all databases in the development set).
    }
    \vspace{-0.2in}
    \label{fig:adapt_curve}
\end{figure}

\paragraph{How much~\modelnameshort~synthesized data should one use for grounded adaptation?}
For this experiment, we vary the amount of cycle-consistent synthesized data used for adaptation.
Figure~\ref{fig:adapt_curve}~shows that that adaptation performance generally increases with the amount of synthesized data in the inference environment, with diminishing return after 30-40k examples.

\section{Related work}

\paragraph{Semantic parsing.}
Semantic parsers parse natural language utterances into executable logical forms with respect to an environment~\citep{zelle1996LearningTP,zettlemoyer2005LearningTM,Liang2011LearningDC}.
In zero-shot semantic parsing, the model is required to generalize to environments (e.g. new domains, new database schemas) not seen during training~\citep{pasupat2015compositional,zhong2017seq2sql,yu2018spider}.
For language-to-SQL zero-shot semantic parsing, a variety of methods have been proposed to generalize to new databases by selecting from table schemas in the new database~\citep{zhang2019editing,guo2019IRNet}.
Our method is complementary to these work --- the synthesis, cycle-consistency, and adaptation steps in~\modelnameshort~can be applied to any parser, so long as we can learn a backward utterance generator and evaluate logical-form equivalence.

\paragraph{Data augmentation.}
Data augmentation transforms original training data to synthesize artificial training data.
\citet{krizhevsky2017imagenet}~crop and rotate input images to improve object recognition.
\citet{dong2017learning}~and~\citet{yu2018qanet}~respectively paraphrase and back-translate~\citep{sennrich2016improving,edunov2018understanding}~questions and documents to improve question-answering.
\citet{jia2016data}~perform data-recombination in the training domain to improve semantic parsing.
\citet{hannun2014deepspeech}~superimpose noisy background tracks with input tracks to improve speech recognition.
Our method is distinct from data-augmentation in the following ways.
First, we synthesize data on logical forms sampled from the new environment instead of the original environment, which allows for adaptation to the new environments.
Second, we propose cycle-consistency to prune low-quality data and keep high-quality data for adaptation.
Our analyses show that these core differences from data-augmentation are central to improving parsing performance.

\paragraph{Cycle-consistent generative adversarial models (cycle-GANs).}
In cycle-GAN~\citep{Zhu2017CycleGAN,Hoffman2017cycada}, a generator forms images that fools a discriminator while the discriminator tries distinguish generated images from naturally occurring images.
The the adversarial objectives of the generator and the discriminator are optimized jointly.
Our method is different from cycle-GANs in that we do not use adversarial objectives and instead rely on matching denotations from executing synthesized queries.
This provides an exact signal compared to potentially incorrect outputs by the discriminator.
Morevoer, cycle-GANs only synthesize the input and verify whether the input is synthesized (e.g.~the utterance looks like a user request).
In contrast,~\modelnameshort~synthesizes both the input and the output, and verifies consistency between the input and the output (e.g.~the utterance matches the query).

\section{Conclusion and Future work}

We proposed~\modelnameshort~to adapt an existing semantic parser to new environments by synthesizing cycle-consistent data.
\modelnameshort~improved parsing performance on three zero-shot parsing tasks.
Our analyses showed that~\modelnameshort~outperforms data augmentation, performance improvement scales with the amount of~\modelnameshort-synthesized data, and cycle-consistency is central to successful adaptation.

In principle,~\modelnameshort~applies to any problems that lack annotated data and differ between training and inference environments.
One such area is robotics, where one trains in simulation because it is prohibitively expensive to collect annotated trajectories in the real world.
In future work, we will consider how to interpret environment specifications to facilitate grounded adaptation in these other areas.

\section*{Acknowledgement}
This research was supported in part by the ARO (W911NF-16-1-0121) and the NSF (IIS-1252835, IIS-1562364).
We thank Julian Michael for helpful discussions, and our reviewers for engaging and constructive feedback.
We also thank Bo Pang and Tao Yu for helping us with running the official evaluations.

\bibliographystyle{acl_natbib}
\bibliography{emnlp2020}

\clearpage
\appendix

\section{Appendix}
\label{sec:appendix}

\subsection{Coverage and multi-turn sampling}
\label{app:sampling}

When we build an empirical distribution over templates on the training set of Spider, we observe a 85\% coverage of dev set templates.
That is, 85\% of dev set examples have a query whose template occurs in the training set.
In other words, while this simple template-filling sampling scheme doesn't provide full coverage over the dev set as a complex grammar would, it covers a large portion of examples.

For Sparc and CoSQL, the sampling procedure is similar to Algorithm~\ref{alg:sample}.
However, because there are two queries (one previous, one current), we first sample a previous query $\coarse^\prime_1$ from $P_{\rm{temp}}(\coarse)$, then sample the current query $\coarse^\prime_2$ from $P_{\rm{temp}}(\coarse | \coarse^\prime_1)$.
As before, the empirical template distributions are obtained by counting templates in the training set.

\subsection{Hyperparameters}
\label{app:hyperparameters}

\begin{table}[ht]
\centering
\begin{tabular}{@{}lccc@{}}
\toprule
\multirow{2}{*}{Dropout location} & \multicolumn{3}{l}{Forward parser} \\ \cmidrule(l){2-4} 
                                  & Spider     & Sparc     & CoSQL     \\ \midrule
post-BERT                         & 0.1        & 0.1       & 0.1       \\
post-enc LSTMs                & 0.1        & 0.3       & 0.1       \\
pre-dec scorer                & 0.1        & 0.1       & 0.3       \\ \bottomrule
\end{tabular}
\caption{Dropout rates for the forward parser.}
\label{tab:dropout-forward}
\end{table}

\begin{table}[ht]
\centering
\begin{tabular}{@{}lccc@{}}
\toprule
\multirow{2}{*}{Dropout location} & \multicolumn{3}{l}{Backward generator} \\ \cmidrule(l){2-4} 
                                  & Spider     & Sparc     & CoSQL     \\ \midrule
post-BERT                     & 0.1        & 0.3       & 0.1       \\
post-enc LSTMs                & 0.1        & 0.1       & 0.1       \\
pre-dec scorer                & 0.1        & 0.1       & 0.3       \\ \bottomrule
\end{tabular}
\caption{Dropout rates for the backward generator.}
\label{tab:dropout-backward}
\end{table}

We use 300-dimensional LSTMs throughout the model.
The BERT model we use is DistilBERT~\citep{sanh2019distilbert}, which we optimize with Adam~\citep{Kingma2014AdamAM} with an initial learning rate of $5e-5$.
We train for 50 epochs with a batch size of 10 and gradient clipping with a norm of 20.
We use dropout after BERT, after encoder LSTMs, and before the pointer scorer.
The values for these dropouts used by our leaderboard submissions are shown in Table~\ref{tab:dropout-forward} and Table~\ref{tab:dropout-backward}.
For each task, these rates are tuned using 3-fold cross-validation with a coarse grid-search over values \{0.1, 0.3\} for each dropout with a fixed seed.

A single training run of the forward parser took approximately 16 hours to run on a single NVIDIA Titan X GPU.
Each task required 3 folds in addition to the final official train/dev run.
For each fold, we grid-searched over dropout rates, which amounts to 8 runs.
In total, we conducted 27 runs on a Slurm cluster.
Including pretrained BERT parameters, the final forward parser contains 142 million parameters.
The final backward utterance generator contains 73 million parameters.

\include{examples}

\subsection{Synthesized examples}
\label{app:synthesized}

In order to quantify the distribution of synthesized examples, we classify synthesized queries according to the difficulty criteria from Spider~\citep{yu2018spider}.
Compared to the Spider development set,~\modelnameshort-synthesized data has an average of 0.60 vs. 0.47 joins, 1.21 vs. 1.37 conditions, 0.20 vs. 0.26 group by’s, 0.23 vs. 0.25 order by’s, 0.07 vs. 0.04 intersections, and 1.25 vs. 1.32 selection columns per query.
This suggests that~\modelnameshort~queries are similar to real data.

Moreover, we example a random sample of 60 synthesized examples.
Out of the 60, 51 are correct.
Mistakes come from aggregation over wrong columns (e.g.~``has the most course'' becomes \texttt{order by sum T2.grade}) and underspecification (e.g.~``lowest of the stadium who has the lowest age'').
There are grammatical errors (e.g.~``that has the most'' becomes ``that has been most''), but most questions are fluent and sensible (e.g.~``find the name and district of the employee that has the highest evaluation bonus'').
A subset of these queries are shown in Table~\ref{tab:synthesized}.

\subsection{Performance breakdown}
\label{app:breakdown}

\begin{table}[ht]
\centering
\addtolength{\tabcolsep}{-1.7pt}
\begin{tabular}{@{}lcccccc@{}}
\toprule
         &    & easy & medium & hard & extra & all  \\ \midrule
count    &    & 470  & 857    & 463  & 357   & 2147 \\ \midrule
baseline & EM & \textbf{75.3} & 54.9   & 45.0 & \textbf{24.8}  & 52.1 \\
         & EX & \textbf{60.3} & 52.7   & 47.5 & 32.6  & 49.8 \\
         & FX & \textbf{73.6} & 52.9   & 44.8 & \textbf{26.4}  & 51.1 \\ \midrule
GAZP     & EM & 73.1 & \textbf{58.7}   & \textbf{47.2} & 23.3  & \textbf{53.3} \\
         & EX & 59.6 & \textbf{59.2}   & \textbf{52.3} & \textbf{33.3}  & \textbf{53.5} \\
         & FX & 71.9 & \textbf{55.3}   & \textbf{46.1} & 24.5  & \textbf{51.7} \\ \bottomrule
\end{tabular}
\caption{Difficulty breakdown for Spider test set.}
\label{tab:breakdown-spider}
\end{table}

\begin{table}[ht]
\centering
\addtolength{\tabcolsep}{-1.7pt}
\begin{tabular}{@{}lcccccc@{}}
\toprule
         &    & easy & medium & hard & extra & all  \\ \midrule
count    &    & 993  & 845    & 399  & 261   & 2498 \\ \midrule
baseline & EM & \textbf{68.9} & 36.9   & 31.2 & 11.1  & 45.9 \\
         & EX & \textbf{61.9} & 35.6   & 30.6 & 18.8  & 43.5 \\
         & FX & \textbf{65.9} & 32.5   & \textbf{28.1} & 10.7  & 42.8 \\ \midrule
GAZP     & EM & 66.5 & \textbf{39.6}   & \textbf{38.4} & \textbf{14.2}  & 45.9 \\
         & EX & 60.1 & \textbf{39.5}   & \textbf{31.1} & \textbf{20.3}  & \textbf{44.6} \\
         & FX & 65.3 & \textbf{36.8}   & 26.3 & \textbf{12.6}  & \textbf{43.9} \\ \bottomrule
\end{tabular}
\caption{Difficulty breakdown for Sparc test set.}
\label{tab:breakdown-sparc}
\end{table}

\begin{table}[ht]
\centering
\addtolength{\tabcolsep}{-1.7pt}
\begin{tabular}{@{}lcccccc@{}}
\toprule
         &    & easy & medium & hard & extra & all  \\ \midrule
count    &    & 730  & 607    & 358  & 209   & 1904 \\ \midrule
baseline & EM & 58.2 & 28.0   & 20.6 & \textbf{18.8}  & 37.2 \\
         & EX & 47.1 & 27.2   & 26.8 & \textbf{28.2}  & 34.9 \\
         & FX & 51.9 & 24.1   & 21.2 & \textbf{20.6}  & 33.8 \\ \midrule
GAZP     & EM & \textbf{60.0} & \textbf{33.8}   & \textbf{23.1} & 13.9  & \textbf{39.7} \\
         & EX & \textbf{48.1} & \textbf{28.3}   & \textbf{41.0} & 23.9  & \textbf{35.9} \\
         & FX & \textbf{55.1} & \textbf{26.9}   & \textbf{25.7} & 16.7  & \textbf{36.3} \\ \bottomrule
\end{tabular}
\caption{Difficulty breakdown for CoSQL test set.}
\label{tab:breakdown-cosql}
\end{table}

\begin{table}[ht]
\centering
\addtolength{\tabcolsep}{-1.7pt}
\begin{tabular}{@{}lccccc@{}}
\toprule
         &    & turn 1 & turn 2 & turn 3 & turn 4+ \\ \midrule
count    &    & 842    & 841    & 613    & 202     \\ \midrule
baseline & EM & \textbf{69.9}   & 41.8   & 28.9   & 16.4    \\
         & EX & \textbf{67.8}   & 36.9   & 28.1   & 16.9    \\
         & FX & \textbf{70.2}   & 35.7   & 24.8   & 13.4    \\ \midrule
GAZP     & EM & 67.8   & 41.9   & \textbf{29.7}   & \textbf{19.6}    \\
         & EX & 66.3   & \textbf{40.1}   & \textbf{29.0}   & \textbf{19.8}    \\
         & FX & 68.8   & \textbf{38.3}   & \textbf{25.9}   & \textbf{18.3}    \\ \bottomrule
\end{tabular}
\caption{Turn breakdown for Sparc test set}
\label{tab:breakdown-sparc-turn}
\end{table}

\begin{table}[ht]
\centering
\addtolength{\tabcolsep}{-1.7pt}
\begin{tabular}{@{}lccccc@{}}
\toprule
         &    & turn 1 & turn 2 & turn 3 & turn 4+ \\ \midrule
count    &    & 548    & 533    & 372    & 351     \\ \midrule
baseline & EM & 47.3   & 36.5   & 32.3   & 28.5    \\
         & EX & 43.8   & \textbf{34.3}   & 30.3   & 27.9    \\
         & FX & 46.2   & 31.9   & 29.4   & 23.4    \\ \bottomrule
GAZP     & EM & \textbf{50.0}   & 36.7   & \textbf{35.7}   & \textbf{30.3}    \\
         & EX & \textbf{46.4}   & 32.3   & \textbf{32.2}   & \textbf{30.2}    \\
         & FX & \textbf{50.0}   & \textbf{32.8}   & \textbf{31.4}   & \textbf{27.1}    \\ \midrule
\end{tabular}
\caption{Turn breakdown for CoSQL test set.}
\label{tab:breakdown-cosql-turn}
\end{table}

In addition to the main experiment results in Table~\ref{tab:results} of Section~\ref{sec:results}, we also examine the performance breakdown across query classes and turns.

\paragraph{GAZP improves performance on harder queries.}
First, we divide queries into difficulty classes following the classification in~\citet{yu2018spider}.
These difficulty classes are based on the number of SQL components, selections, and conditions.
For example, queries that contain more SQL keywords such as \texttt{GROUP BY}, \texttt{ORDER BY}, \texttt{INTERSECT}, nested subqueries, column selections, and aggregators, etc are considered to be harder.
\citet{yu2018spider} shows examples of SQL queries in the four hardness categories.
Note that \textbf{extra} is a catch-all category for queries that exceed qualifications of \textbf{hard}, as a result it includes artifacts (e.g.~set exclusion operations) that may introduce other confounding factors.
Tables~\ref{tab:breakdown-spider},~\ref{tab:breakdown-sparc}, and~\ref{tab:breakdown-cosql}~ respectively break down the performance of models on Spider, Sparc, and CoSQL.
We observe that the gains in~\modelnameshort~are generally more pronounced in more difficult queries.
This finding is consistent across tasks (with some variance) and across three evaluation metrics.

One potential explanation for this gain is that the generalization problem is exacerbated in more difficult queries.
Consider the example of language-to-SQL parsing, in which we have trained a parser on an university database and are now evaluating it on a sales database.
While it is difficult to produce simple queries in the sales database due to ta lack of training data, it is likely even more difficult to produce nested queries, queries with groupings, queries with multiple conditions, etc.
Because~\modelnameshort~synthesizes queries --- including difficult ones --- in the sales database, the adapted parser learns to handle these cases.
In contrast, simpler queries are likely easier to learn, hence adaptation does not help as much.

\paragraph{GAZP improves performance in longer interactions.}
For Sparc and CoSQL, which include multi-turn interactions between the user and the system, we divide queries into how many turns into the interaction they occur.
This classification in described in~\citet{yu2019sparc}~and~\citet{yu2019cosql}.
Tables~\ref{tab:breakdown-sparc-turn}~and~\ref{tab:breakdown-cosql-turn}~ respectively break down the performance of models on Sparc and CoSQL.
We observe that the gains in~\modelnameshort~are more pronounced in turns later in the interaction.
Against, this finding is consistent not only across tasks, but across the three evaluation metrics.

A possible reason for this gain is that the conditional sampling procedure shown in Algorithm~\ref{alg:sample}~improves multi-turn parsing by synthesizing multi-turn examples.
How much additional variation should we expect in a multi-turn setting?
Suppose we discover $T$ coarse-grain templates by counting the training data, where each coarse-grain template has $S$ slots on average.
For simplicity, let us ignore value slots and only consider column slots.
Given a new database with $N$ columns, the number of possible filled queries is on the order of $O\left(T \times \binom SN\right)$.
For $K$ turns, the number of possible queries sequences is then $O\left(\left(T \times \binom SN\right)^K\right)$.
This exponential increase in query variety may improve parser performance on later-turn queries (e.g. those with a previous interaction), which in turn reduce cascading errors throughout the interaction.

\end{document}

%% file: victor.tex
\newcommand{\tocite}[1]{{[\hl{CITE: #1]}}}
\newcommand{\todo}[1]{{[\hl{TODO: #1}]}}
\newcommand{\victor}[1]{{[\hl{VICTOR: #1}]}}
\newcommand{\luke}[1]{{[\hl{LUKE: #1}]}}
\newcommand{\mike}[1]{{[\hl{MIKE: #1}]}}
\newcommand{\sida}[1]{{[\hl{SIDA: #1}]}}

\newcommand{\papertitle}{{Grounded Adaptation for Zero-shot Executable Semantic Parsing}}
\newcommand{\modelnameemphasized}{{\textbf{G}rounded \textbf{A}daptation for \textbf{Z}ero-shot Executable Semantic \textbf{P}arsing}}
\newcommand{\modelnameshort}{{GAZP}}

\newcommand{\caveat}{{$^*$}}
\newcommand{\sfunction}[1]{\textsf{\textsc{#1}}}
\newcommand{\logicalform}{{q}}
\newcommand{\glogicalform}{{\logicalform^\prime}}
\newcommand{\denotation}{{\rm ans}}
\newcommand{\gdenotation}{{\denotation^\prime}}
\newcommand{\utterance}{{u}}
\newcommand{\gutterance}{{\utterance^\prime}}
\newcommand{\environment}{{e}}
\newcommand{\infenvironment}{{\environment^\prime}}
\newcommand{\parser}{{F}}
\newcommand{\generator}{{G}}
\newcommand{\exe}[2]{{\sfunction{Exe}(#1, #2)}}
\newcommand{\tokutterance}{{w_{\utterance}}}
\newcommand{\tokenvironment}{{w_{\environment}}}
\newcommand{\bertforward}{{{\rm BERT}_{\rightarrow}}}
\newcommand{\bertbackward}{{{\rm BERT}_{\leftarrow}}}

\newcommand{\forward}[1]{{\overrightarrow{#1}}}
\newcommand{\backward}[1]{{\overleftarrow{#1}}}

\newcommand{\state}{{h}}
\newcommand{\enc}{{\rm enc}}
\newcommand{\dec}{{\rm dec}}
\newcommand{\bertenc}{{B}}
\newcommand{\forwardbertenc}{{\forward{\bertenc}}}
\newcommand{\backwardbertenc}{{\backward{\bertenc}}}
\newcommand{\emb}{{\rm emb}}
\newcommand{\txt}{{d}}
\newcommand{\phrase}{{x}}
\newcommand{\lstm}{{\rm LSTM}}
\newcommand{\softmax}{{\rm softmax}}
\newcommand{\concat}[1]{{[#1]}}
\newcommand{\rep}[1]{{{\rm rep} (#1)}}
\newcommand{\cand}{{c}}
\newcommand{\syn}{{s}}
\newcommand{\vocab}{{v}}
\newcommand{\template}{{\hat{\logicalform}}}
\newcommand{\coarse}{{z}}
\newcommand{\Coarse}{{Z}}
\newcommand{\envdesc}{{w}}
\newcommand{\envdoc}{{d}}
\newcommand{\auglogicalform}{{\tilde{\logicalform}}}

\newcommand{\forwardcand}{{\forward{c}}}
\newcommand{\backwardcand}{{\backward{c}}}
\newcommand{\argmax}{{{\rm argmax}}}
\newcommand*{\Comb}[2]{{}^{#1}C_{#2}}%

\newcommand{\real}[1]{{\mathbb{R}^{{#1}}}}

\newcommand{\weight}{\bm W}
\newcommand{\vectorweight}{\bm w}
\newcommand{\bias}{\bm b}
\newcommand{\scalarbias}{b}
\newcommand{\vectorgamma}{\bm \gamma}
\newcommand{\vectorbeta}{\bm \beta}
\newcommand{\matrixgamma}{\bm \Gamma}
\newcommand{\matrixbeta}{\bm \Beta}
\newcommand{\dimm}{d}
\newcommand{\len}{l}
\newcommand{\filmfeat}{\bm x}
\newcommand{\filmmatrixfeat}{\bm X}
\newcommand{\filmout}{\bm V}
\newcommand{\filmsumm}{\bm s}
\newcommand{\rnnhid}{\bm H}
\newcommand{\vectorrnnhid}{\bm h}
\newcommand{\attnscore}{\bm a^\prime}
\newcommand{\scalarattnscore}{a^\prime}
\newcommand{\normattnscore}{\bm a}
\newcommand{\scalarnormattnscore}{a}
\newcommand{\attncontext}{{\bm c}}
\newcommand{\outrep}{\filmsumm^{(\mathrm{last})}}
\newcommand{\outtransrep}{\bm o}

\newcommand{\policy}{{\bm y}_{{\rm policy}}}
\newcommand{\baseline}{y_{\rm baseline}}

\newcommand{\selfattn}{{\rm selfattn}}
\newcommand{\BiLSTM}{{\rm BiLSTM}}
\newcommand{\mlp}{{\rm MLP}}
\newcommand{\conv}{{\rm Conv}}
\newcommand{\relu}{{\rm ReLU}}
\newcommand{\maxpool}{{\rm MaxPool}}
\newcommand{\attend}[2]{{\rm attend}({{#1}}, {{#2}})}

%% file: examples.tex
\begin{table*}[t]
    \centering
    \begin{tabular}{p{0.3\linewidth} p{0.65\linewidth}}
    \toprule
    list all the last name of owners in alphabetical order .
    &
    \texttt{select last\_name from Owners order by last\_name}
    \\ \midrule
    how many friend are there ?
    &
    \texttt{select count ( * ) from Friend}
    \\ \midrule
    what is the id of the votes that has been most distinct contestants ?
    &
    \texttt{"select T2.vote\_id from CONTESTANTS as T1 join VOTES as T2 on T1.contestant\_number = T2.contestant\_number group by ( T2.vote\_id ) order by count ( T1.contestant\_number ) desc limit 1}
    \\ \midrule
    what are the name of higher ?
    &
    \texttt{select name from Highschooler}
    \\ \midrule
    how many car makers has the horsepower of 81 ?
    &
    \texttt{select count ( * ) from cars\_data as T1 join car\_names as T2 on T1.Id = T2.MakeId join model\_list as T3 on T2.Model = T3.Model join car\_makers as T4 on T3.Maker = T4.Id where T1.Horsepower = '81'}
    \\ \midrule
    what are the starts of hiring who are located in the city of Bristol ?
    &
    \texttt{select T2.Start\_from from employee as T1 join hiring as T2 on T1.Employee\_ID = T2.Employee\_ID where T1.City = 'Bristol'}
    \\ \midrule
    find the name and district of the employee that has the highest evaluation bonus .
    &
    \texttt{select T2.Name , T4.District from evaluation as T1 join employee as T2 on T1.Employee\_ID = T2.Employee\_ID join hiring as T3 on T2.Employee\_ID = T3.Employee\_ID join shop as T4 on T3.Shop\_ID = T4.Shop\_ID order by T1.Bonus desc limit 1}
    \\ \midrule
    what is the cell number of the owners with the largest charges amount ?
    &
    \texttt{select T1.cell\_number from Owners as T1 join Charges as T2 order by T2.charge\_amount desc limit 1}
    \\ \midrule
    what is the minimum , average , and maximum grade of all high schooler ?
    &
    \texttt{select min ( grade ) , avg ( grade ) , max ( grade ) from Highschooler}
    \\ \midrule
    what is the age of the teacher who has the most course ?
    &
    \texttt{select T1.Age from teacher as T1 join course\_arrange as T2 on T1.Teacher\_ID = T2.Teacher\_ID group by T2.Teacher\_ID order by sum ( T2.Grade ) desc limit 1}
    \\ \bottomrule
    \end{tabular}
    \caption{Examples of synthesized queries}
    \label{tab:synthesized}
\end{table*}

%% file: emnlp2020.bbl
\begin{thebibliography}{30}
\expandafter\ifx\csname natexlab\endcsname\relax\def\natexlab#1{#1}\fi

\bibitem[{Bahdanau et~al.(2015)Bahdanau, Cho, and
  Bengio}]{bahdanau-align_and_translate}
Dzmitry Bahdanau, Kyunghyun Cho, and Yoshua Bengio. 2015.
\newblock Neural machine translation by jointly learning to align and
  translate.
\newblock \emph{ICLR}.

\bibitem[{Cao et~al.(2019)Cao, Zhu, Liu, Li, and Yu}]{cao2019dual}
Ruisheng Cao, Su~Zhu, Chen Liu, Jieyu Li, and Kai Yu. 2019.
\newblock Semantic parsing with dual learning.
\newblock In \emph{ACL}.

\bibitem[{Devlin et~al.(2019)Devlin, Chang, Lee, and
  Toutanova}]{devlin2018bert}
Jacob Devlin, Ming-Wei Chang, Kenton Lee, and Kristina Toutanova. 2019.
\newblock Bert: Pre-training of deep bidirectional transformers for language
  understanding.
\newblock In \emph{NAACL}.

\bibitem[{Dong et~al.(2017)Dong, Mallinson, Reddy, and
  Lapata}]{dong2017learning}
Li~Dong, Jonathan Mallinson, Siva Reddy, and Mirella Lapata. 2017.
\newblock Learning to paraphrase for question answering.
\newblock In \emph{EMNLP}.

\bibitem[{Edunov et~al.(2018)Edunov, Ott, Auli, and
  Grangier}]{edunov2018understanding}
Sergey Edunov, Myle Ott, Michael Auli, and David Grangier. 2018.
\newblock Understanding back-translation at scale.
\newblock In \emph{EMNLP}.

\bibitem[{Fried et~al.(2018)Fried, Hu, Cirik, Rohrbach, Andreas, Morency,
  Berg-Kirkpatrick, Saenko, Klein, and Darrell}]{fried2018speaker}
Daniel Fried, Ronghang Hu, Volkan Cirik, Anna Rohrbach, Jacob Andreas,
  Louis-Philippe Morency, Taylor Berg-Kirkpatrick, Kate Saenko, Dan Klein, and
  Trevor Darrell. 2018.
\newblock Speaker-follower models for vision-and-language navigation.
\newblock In \emph{NeurIPS}.

\bibitem[{Graesser et~al.(2005)Graesser, Chipman, Haynes, and
  Olney}]{graesser2005autotutor}
Arthur~C Graesser, Patrick Chipman, Brian~C Haynes, and Andrew Olney. 2005.
\newblock {AutoTutor}: An intelligent tutoring system with mixed-initiative
  dialogue.
\newblock \emph{IEEE Transactions on Education}.

\bibitem[{Guo et~al.(2019)Guo, Zhan, Gao, Xiao, Lou, Liu, and
  Zhang}]{guo2019IRNet}
Jiaqi Guo, Zecheng Zhan, Yan Gao, Yan Xiao, Jian{-}Guang Lou, Ting Liu, and
  Dongmei Zhang. 2019.
\newblock Towards complex text-to-{SQL} in cross-domain database with
  intermediate representation.
\newblock In \emph{ACL}.

\bibitem[{Hannun et~al.(2014)Hannun, Case, Casper, Catanzaro, Diamos, Elsen,
  Prenger, Satheesh, Sengupta, Coates, and Ng}]{hannun2014deepspeech}
Awni~Y. Hannun, Carl Case, Jared Casper, Bryan Catanzaro, Greg Diamos, Erich
  Elsen, Ryan Prenger, Sanjeev Satheesh, Shubho Sengupta, Adam Coates, and
  Andrew~Y. Ng. 2014.
\newblock {Deep Speech}: Scaling up end-to-end speech recognition.
\newblock \emph{CoRR}, abs/1412.5567.

\bibitem[{Hochreiter and Schmidhuber(1997)}]{Hochreiter1997LongSM}
Sepp Hochreiter and J{\"u}rgen Schmidhuber. 1997.
\newblock Long short-term memory.
\newblock \emph{Neural computation}.

\bibitem[{Hoffman et~al.(2018)Hoffman, Tzeng, Park, Zhu, Isola, Saenko, Efros,
  and Darrell}]{Hoffman2017cycada}
Judy Hoffman, Eric Tzeng, Taesung Park, Jun-Yan Zhu, Phillip Isola, Kate
  Saenko, Alexei~A. Efros, and Trevor Darrell. 2018.
\newblock {CyCADA}: Cycle consistent adversarial domain adaptation.
\newblock In \emph{ICML}.

\bibitem[{Jia and Liang(2016)}]{jia2016data}
Robin Jia and Percy Liang. 2016.
\newblock Data recombination for neural semantic parsing.
\newblock In \emph{ACL}.

\bibitem[{Kingma and Ba(2015)}]{Kingma2014AdamAM}
Diederik~P. Kingma and Jimmy Ba. 2015.
\newblock Adam: A method for stochastic optimization.
\newblock In \emph{ICLR}.

\bibitem[{Krizhevsky et~al.(2017)Krizhevsky, Sutskever, and
  Hinton}]{krizhevsky2017imagenet}
Alex Krizhevsky, Ilya Sutskever, and Geoffrey~E. Hinton. 2017.
\newblock Imagenet classification with deep convolutional neural networks.
\newblock \emph{Communications of the ACM}.

\bibitem[{Liang et~al.(2011)Liang, Jordan, and Klein}]{Liang2011LearningDC}
Percy Liang, Michael~I. Jordan, and Dan Klein. 2011.
\newblock Learning dependency-based compositional semantics.
\newblock \emph{Computational Linguistics}.

\bibitem[{Matuszek et~al.(2013)Matuszek, Herbst, Zettlemoyer, and
  Fox}]{matuszek2013Learning}
Cynthia Matuszek, Evan Herbst, Luke Zettlemoyer, and Dieter Fox. 2013.
\newblock Learning to parse natural language commands to a robot control
  system.
\newblock \emph{Experimental Robotics}.

\bibitem[{Pasupat and Liang(2015)}]{pasupat2015compositional}
Panupong Pasupat and Percy Liang. 2015.
\newblock Compositional semantic parsing on semi-structured tables.
\newblock In \emph{ACL}.

\bibitem[{Sanh et~al.(2020)Sanh, Debut, Chaumond, and
  Wolf}]{sanh2019distilbert}
Victor Sanh, Lysandre Debut, Julien Chaumond, and Thomas Wolf. 2020.
\newblock Distilbert, a distilled version of bert: smaller, faster, cheaper and
  lighter.
\newblock In \emph{NeurIPS Workshop on Energy Efficient Machine Learning and
  Cognitive Computing}.

\bibitem[{Sennrich et~al.(2016)Sennrich, Haddow, and
  Birch}]{sennrich2016improving}
Rico Sennrich, Barry Haddow, and Alexandra Birch. 2016.
\newblock Improving neural machine translation models with monolingual data.
\newblock In \emph{ACL}.

\bibitem[{Vinyals et~al.(2015)Vinyals, Fortunato, and
  Jaitly}]{vinyals-pointer_networks}
Oriol Vinyals, Meire Fortunato, and Navdeep Jaitly. 2015.
\newblock Pointer networks.
\newblock In \emph{NIPS}.

\bibitem[{Yih et~al.(2014)Yih, He, and Meek}]{yih2014semantic}
Wen-tau Yih, Xiaodong He, and Christopher Meek. 2014.
\newblock Semantic parsing for single-relation question answering.
\newblock In \emph{ACL}.

\bibitem[{Yu et~al.(2018{\natexlab{a}})Yu, Dohan, Luong, Zhao, Chen, Norouzi,
  and Le}]{yu2018qanet}
Adams~Wei Yu, David Dohan, Minh-Thang Luong, Rui Zhao, Kai Chen, Mohammad
  Norouzi, and Quoc~V Le. 2018{\natexlab{a}}.
\newblock Qanet: Combining local convolution with global self-attention for
  reading comprehension.
\newblock In \emph{ICLR}.

\bibitem[{Yu et~al.(2019{\natexlab{a}})Yu, Zhang, Er, Li, Xue, Pang, Lin, Tan,
  Shi, Li, Jiang, Yasunaga, Shim, Chen, Fabbri, Li, Chen, Zhang, Dixit, Zhang,
  Xiong, Socher, Lasecki, and Radev}]{yu2019cosql}
Tao Yu, Rui Zhang, He~Yang Er, Suyi Li, Eric Xue, Bo~Pang, Xi~Victoria Lin,
  Yi~Chern Tan, Tianze Shi, Zihan Li, Youxuan Jiang, Michihiro Yasunaga,
  Sungrok Shim, Tao Chen, Alexander Fabbri, Zifan Li, Luyao Chen, Yuwen Zhang,
  Shreya Dixit, Vincent Zhang, Caiming Xiong, Richard Socher, Walter Lasecki,
  and Dragomir Radev. 2019{\natexlab{a}}.
\newblock Cosql: A conversational text-to-sql challenge towards cross-domain
  natural language interfaces to databases.
\newblock In \emph{EMNLP}.

\bibitem[{Yu et~al.(2018{\natexlab{b}})Yu, Zhang, Yang, Yasunaga, Wang, Li, Ma,
  Li, Yao, Roman, Zhang, and Radev}]{yu2018spider}
Tao Yu, Rui Zhang, Kai Yang, Michihiro Yasunaga, Dongxu Wang, Zifan Li, James
  Ma, Irene Li, Qingning Yao, Shanelle Roman, Zilin Zhang, and Dragomir Radev.
  2018{\natexlab{b}}.
\newblock Spider: A large-scale human-labeled dataset for complex and
  cross-domain semantic parsing and text-to-sql task.
\newblock In \emph{EMNLP}.

\bibitem[{Yu et~al.(2019{\natexlab{b}})Yu, Zhang, Yasunaga, Tan, Lin, Li, Er,
  Li, Pang, Chen, Ji, Dixit, Proctor, Shim, Kraft, Zhang, Xiong, Socher, and
  Radev}]{yu2019sparc}
Tao Yu, Rui Zhang, Michihiro Yasunaga, Yi~Chern Tan, Xi~Victoria Lin, Suyi Li,
  Heyang Er, Irene Li, Bo~Pang, Tao Chen, Emily Ji, Shreya Dixit, David
  Proctor, Sungrok Shim, Jonathan Kraft, Vincent Zhang, Caiming Xiong, Richard
  Socher, and Dragomir Radev. 2019{\natexlab{b}}.
\newblock Sparc: Cross-domain semantic parsing in context.
\newblock In \emph{ACL}.

\bibitem[{Zelle and Mooney(1996)}]{zelle1996LearningTP}
John~M. Zelle and Raymond~J. Mooney. 1996.
\newblock Learning to parse database queries using inductive logic programming.
\newblock In \emph{AAAI/IAAI}.

\bibitem[{Zettlemoyer and Collins(2005)}]{zettlemoyer2005LearningTM}
Luke~S. Zettlemoyer and Michael Collins. 2005.
\newblock Learning to map sentences to logical form: Structured classification
  with probabilistic categorial grammars.
\newblock In \emph{UAI}.

\bibitem[{Zhang et~al.(2019)Zhang, Yu, Er, Shim, Xue, Lin, Shi, Xiong, Socher,
  and Radev}]{zhang2019editing}
Rui Zhang, Tao Yu, He~Yang Er, Sungrok Shim, Eric Xue, Xi~Victoria Lin, Tianze
  Shi, Caiming Xiong, Richard Socher, and Dragomir Radev. 2019.
\newblock Editing-based sql query generation for cross-domain context-dependent
  questions.
\newblock In \emph{EMNLP}.

\bibitem[{Zhong et~al.(2017)Zhong, Xiong, and Socher}]{zhong2017seq2sql}
Victor Zhong, Caiming Xiong, and Richard Socher. 2017.
\newblock {Seq2SQL}: Generating structured queries from natural language using
  reinforcement learning.
\newblock \emph{CoRR}, abs/1709.00103.

\bibitem[{Zhu et~al.(2017)Zhu, Park, Isola, and Efros}]{Zhu2017CycleGAN}
Jun-Yan Zhu, Taesung Park, Phillip Isola, and Alexei~A Efros. 2017.
\newblock Unpaired image-to-image translation using cycle-consistent
  adversarial networks.
\newblock In \emph{ICCV}.

\end{thebibliography}
